\documentclass[conference]{IEEEtran_temp}
\makeatletter

\def\small{\@setfontsize{\small}{8}{10pt}}
\def\ps@IEEEtitlepagestyle{%
    \def\@oddfoot{\mycopyrightnotice}%
    \def\@evenfoot{}%
}
\def\mycopyrightnotice{%
    {\footnotesize  }
    \gdef\mycopyrightnotice{}
}



\makeatletter
\newcommand*\titleheader[1]{\gdef\@titleheader{#1}}
\AtBeginDocument{%
  \let\st@red@title\@title%
  \def\@title{%
    \bgroup\normalfont\small\raggedleft\@titleheader\par\egroup
    \vskip1.5em\st@red@title}
}
\makeatother

\ifCLASSINFOpdf
\else
\fi
\usepackage{amsmath,amsfonts,amssymb}
\usepackage{bm}
\usepackage{subfigure}
\usepackage{url}
\usepackage{longtable}
\usepackage[T1]{fontenc}
\usepackage{subfig}
\usepackage{graphicx}
\usepackage{graphics}
\usepackage[space]{cite}

\usepackage{algorithm}
\usepackage{algorithmic}
\usepackage[flushleft]{threeparttable}

\usepackage{mathtools,nccmath}

\title{Sparse Gaussian Process Based On Hat Basis Functions}

\titleheader{\textit{}}
\begin{document}

%
%


\author{\IEEEauthorblockN{Wenqi Fang, Huiyun Li, Hui Huang, Shaobo Dang, Zhejun Huang, Zheng Wang}
\IEEEauthorblockA{
Shenzhen Institutes of Advanced Technology, Chinese Academy of Sciences\\
Shenzhen, P.R.China\\
Email: wq.fang@siat.ac.cn}
}

\maketitle

\begin{abstract}
Gaussian process is one of the most popular non-parametric Bayesian methodologies for modeling the regression problem. 
It is completely determined by its mean and covariance functions. And its linear property makes it relatively straightforward to solve the prediction problem.
Although Gaussian process has been successfully applied in many fields, it is still not enough to deal with physical systems that satisfy inequality constraints. 
This issue has been addressed by the so-called constrained Gaussian process in recent years. In this paper, we extend the core ideas of constrained Gaussian process. 
According to the range of training or test data, we redefine the hat basis functions mentioned in the constrained Gaussian process. Based on hat basis functions, we propose a new sparse Gaussian process method to solve the unconstrained regression problem. Similar to the exact Gaussian process and Gaussian process with Fully Independent Training Conditional approximation, 
our method obtains satisfactory approximate results on open-source datasets or analytical functions. 
In terms of performance, the proposed method reduces the overall computational complexity from $O(n^{3})$ computation in exact Gaussian process to $O(nm^{2})$ with $m$ hat basis functions and $n$ training data points.

\end{abstract}

\begin{IEEEkeywords}
Gaussian process; hat basis function; sparse Gaussian process; spectral approximation

\end{IEEEkeywords}

%
\IEEEpeerreviewmaketitle

\section{Introduction}
In recent years, Gaussian process (GP) regression has become one of the prevailing regression techniques\cite{RW}. To be precise, a GP is a distribution over functions such that any  finite set of function values have a joint Gaussian distribution. It enjoys analytical tractability, and has a fully probabilistic work-flow. The obtained mean function and covariance function are used for regression and uncertainty estimation, respectively. The strength of GP regression lies in avoiding overfitting while still finding functions complex enough to describe any observed behaviors, even in noisy or unstructured data. GP is usually applied to the cases when observations are rare or expensive to produce and methods such as deep learning performs poorly. GP has been applied among a wide range from robotics\cite{cardelli2019robustness}, biology\cite{overstall2020bayesian}, global optimization\cite{yang2019multi}, astrophysics\cite{iyer2019nonparametric} to engineering\cite{jiang2019review}. 

However, direct implemented GP  has the computational and memory requirement in order of  $O(n^3)$ and $O(n^2)$, where $n$ is the number of training data, which limits its usage for application in current big data era. To overcome this problem, over the years, several sparse GP schemes have been proposed. Currently, there are mainly three categories of sparsity-related methods\cite{solin2020hilbert}: inducing input methods based on the Nystr\"om approximation\cite{bui2017unifying}, direct spectral approximations\cite{cramer2013stationary} and structure exploiting techniques\cite{wilson2015kernel}. In addition, the output of GP may fail to meet the specific physical requirement. In many situations, the physical system may be known to satisfy inequality constraints with respect to some input or output variables. Consequently, introducing inequality constraints in GP models can lead to more realistic uncertainties in learning a great variety of real-world problems. The research on this issue has given birth to an emerging research area named as constrained GP\cite{maatouk2017gaussian,lopez2018finite}.

In this paper, we redefine the hat basis functions in the arbitrary domain and put forward a novel sparse GP method, which reduces the overall computational complexity from $O(n^{3})$ to $O(nm^{2})$, inspired by the core ideas of constrained GP. The proposed method is exactly kind of direct spectral approximations, and it can be adopted to solve the regression problem, which does not subject to inequality constraints, by optimization method\cite{RW}. We demonstrate that our method, named as Hat-GP, achieves comparable results to the existing state-of-the-art methods.

The rest of the paper is organized as follows. In Sec. \textrm{II}, we present a brief introduction to the GP and GP with Fully Independent Training Conditional (FITC) Approximation. In Sec. \textrm{III}, we illustrate our methods thoroughly. In Sec. \textrm{III}, we benchmark our results to some existing methods by through numerical experiments. Finally, we conduct some necessary discussions and conclude our work in Sec. \textrm{IV}.

\section{Backgound}
We briefly review GP regression and GP regression with FITC approximation. For more details on Gaussian processes, we refer the reader to the book written by Rasmussen and Williams\cite{RW}.
\subsection{Gaussian process}
GP prior over function $f(x)$ implies that any set of function values
$f$, indexed by the input $x$, have a multivariate
Gaussian distribution
\begin{equation}\label{eq_GP_prior}
p(f|x) = N(f|\mu(x), K(x,x)),
\end{equation}
where $K$ is the covariance matrix and $\mu$ is non-zero mean value.  
In our paper, we set the mean value as a constant $\mu_c$. 
It need to be augmented into a matrix according to the variables to be solved in the numerical simulation.
The covariance matrix characterizes the correlation between different points in this stochastic process. 

Typically, it is more realistic that in our training data, only the noisy observations are available rather than the precise function value of $f(x)$. 
We assume that the noise is additive, independent and Gaussian, such that the
relationship between the function $f(x)$ and the observed noisy targets $y$ are given by:
\begin{equation}\label{eq_GP_prior}
y = f(x)+\varepsilon,
\end{equation}
where $\varepsilon$ $\sim$ $N(0, \sigma^2 I)$ and $I$ is identity matrix.  In this situation, the mean and variance of exact posterior distribution $p(y^{*}|x^{*}, x, y)$ with 
test input $x^{*}$ is as follows:
\begin{align}
&\mu_c+K(x^{*}, x)(K(x, x)+\sigma^2I)^{-1}(y-\mu_c)     \\
&K(x^{*}, x^{*})+\sigma^2I - K(x^{*}, x)(K(x, x)+\sigma^2I)^{-1} K(x, x^{*})).
\end{align}
Covariance function can be chosen freely as long as the produced
covariance matrices are symmetric and positive semi-definite. A
common stationary covariance function is the squared exponential (SE) kernel
\begin{equation}
K(x_i,x_j)=\sigma_{se}^2\exp(- \frac{(x_{i}-x_{j})^2}{2 l^2}),
\end{equation}
where $\theta=[\sigma_{se}, l]$ are the hyper-parameters. $\sigma_{se}$ is the scaling parameter, and $l$ is the length-scale parameter, governing
how fast the correlation decreases as the distance increases. For the multi-dimensional input, a flexible way to model functions is to multiply together kernels defined on each individual dimension. For example,  a product of squared exponential kernels over $d$-dimensional inputs is, each having a different length-scale parameter, 
\begin{equation}
K(x_i,x_j)=\sigma_{se}^2\exp(-\sum_{k=1}^d\frac{ (x_{i,k}-x_{j,k})^2}{2 l_k^2}).
\end{equation}
This is often called automatic relevance determination (ARD) SE kernel, 
so named because the length-scale parameters $l_1$, $l_2$, . . ., $l_d$, implicitly determines the relevance of each dimension. 
The input dimensions with relatively large length-scales imply relatively little variation along those dimensions in the function being modeled. 
Other common used kernel functions are discussed in\cite{duvenaud2014automatic}.

To use GP to solve the problem, we need to infer the parameters in the model during 
training procedure. A crucial character for GP is that we can calculate its marginal likelihood, 
 and it can help significantly for model selection. A popular way to tune kernel parameters is to 
 maximize the log marginal likelihood. The marginal likelihood is, given the parameters $
\sigma$ and $\theta$,
$p(y| x, \theta,\sigma) = \int
p(y|f,\sigma)p(f|x,\theta) df$.
With a Gaussian likelihood, it has an analytic closed-form which gives the negative log
marginal likelihood:
\begin{equation}\label{eq_log_marginal_likelihood}
\begin{aligned}
-\log p(y|x, \theta,\sigma) &=  \frac{n}{2}\log(2\pi) +\frac{1}{2}\log |K(x, x) + \sigma^2I|    \\
&+\frac{1}{2} (y-\mu_c)^{\text{T}} (K(x, x) + \sigma^2I)^{-1}(y- \mu_c),
\end{aligned}
\end{equation}
where $n$ is the number of the input data and $\left| \cdot  \right| $ represents 
determinant of the matrix. For convenience, we call this method as Exact-GP.
If the likelihood is not Gaussian, the marginal likelihood needs to be approximated. 
Many approximate methods can be used, like Laplace approximation\cite{hartmann2019laplace}, variational method\cite{zhang2018advances} and sampling method\cite{benton2019function}.

\subsection{Gaussian process with FITC Approximation}
GP method with FITC approximation, here we call it FITC-GP for short, is a wildly adopted sparse GP method based on the Nystr\"om approximation. 
It was originally called sparse Gaussian Processes using pseudo-inputs (SGPP) which was proposed by Snelson and Ghahraman\cite{snelson2006sparse}. It was later reformulated by Quinonero-Candela and Rasmussen\cite{quinonero2005unifying, naish2008generalized}. FITC-GP does not form the full covariance matrix over all $n$ training inputs. Instead, it introduces $m$ ($m$$<$$n$) pseudo points $u$ as an auxiliary set, and their positions can be optimized throughout the domain to improve accuracy. It reduces the computation complexity of Exact-GP from $O(n^{3})$ to $O(nm^{2})$. 
Accoring to Quinonero-Candela's paper, it is shown that the FITC-GP can be considered not just as an approximation, and it is exactly equivalent to modify the GP prior as following:
\begin{equation}\label{Qff}
p(f|x) = N(\mu_c, K_{FITC}(x, x))
\end{equation}
where $\mu_c$ is its constant mean and  the covariance matrix $K_{FITC}$ is: $K_{FITC} = Q_{ff} + diag[K_{ff} - Q_{ff} ])$. 
The matrix $Q_{ff}$ is the Nystr\"om approximation to the full covariance matrix: 
\begin{equation}\label{Qff}
Q_{ff} = K_{fu}K^{-1}_{uu}K_{uf}.
\end{equation}
And $K_{**}$ ($*$ stands for $u$ or $f$) is the evaluation of the covariance function at pseudo inputs, training data $x$ or between them. 
It follows the same principles as Exact-GP. As for the log marginal likelihood in FITC-GP, its mathematical expression is the same as  
Exact-GP except for replacing the kernel with $K_{FITC}$. The hyperparameters in the FITC-GP model and inducing inputs 
can be trained jointly by optimizing its log marginal likelihood\cite{snelson2006sparse, naish2008generalized}.

\section{Hat-GP method for regression problem}
In this section, we introduce the main property of hat basis functions and present our detailed algorithm to approximate GP regression. 
\subsection{Hat basis function}
\begin{figure}[!t]
\centering
\includegraphics[width=3in]{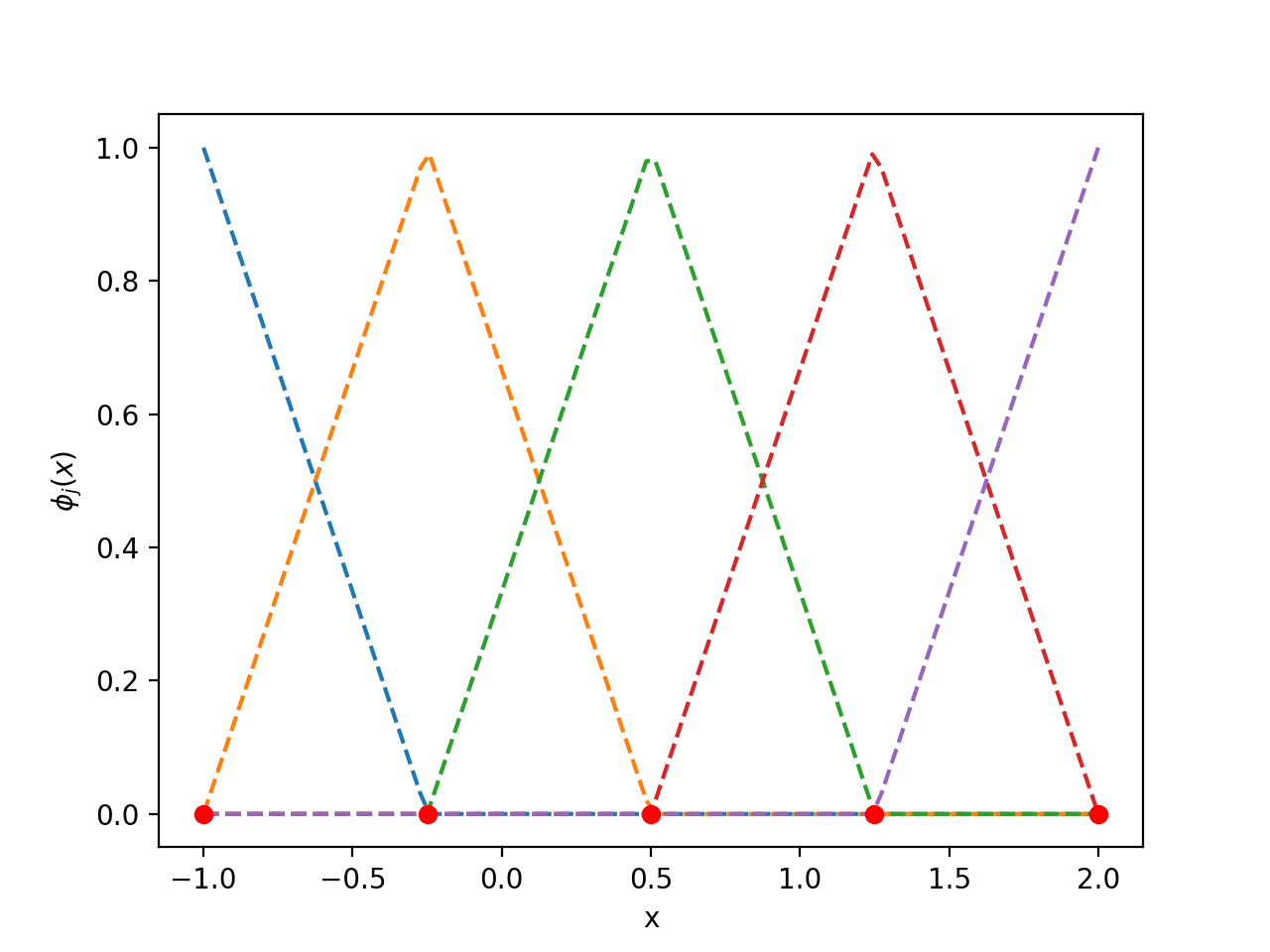}
\caption{Hat basis functions defined in domain $[-1.0, 2.0]$, and the red dots in the figure are the defined knots $t_j$.}
\label{hat}
\end{figure}

Let $Y \in R$ be a GP with fixed mean $\mu_c$ and covariance function $K$ (the kernel parameters denoted as $\theta$). In other papers, the hat function is usually defined in the domain $[0, 1]$\cite{maatouk2017gaussian, lopez2018finite}. However, this definition will make the final predicted results which are outside of domain $[0, 1]$  collapse to zero. To avoid this issue, we consider $x \in \mathcal{D}$ with compact input space $\mathcal{D} = [lb, ub]$, where $lb$ and $ub$ are defined as floor and ceiling of minimum and maximum of training data $x$ or test data $x^{*}$, and a set of knots $t_0, \cdots, t_{m-1}$. In this way, we don't need to regulate the input space to be in domain $[0,1]$. For simplicity, we consider equally-spaced knots $t_j = lb + j \Delta_m$ with $\Delta_m = (ub - lb)/(m-1)$ in this paper. Then, we define a finite-dimensional GP, denoted by $Y_m$, as the piecewise linear interpolation of $Y$ at knots $t_0, \cdots, t_{m-1}$:
\begin{align}
Y_m (x) = \sum_{j=1}^{m} Y(t_j) \phi_j (x),  \quad \mbox{s.t.} \quad Y_m(x_i) + \varepsilon_i \approx y_i
\label{eq1}
\end{align}
where $\phi_0, \cdots, \phi_{m-1}$ are hat basis functions given by
\begin{align}
\phi_j (x) :=
\begin{cases}
1 - \left|\frac{x - t_j}{\Delta_m}\right| & \mbox{if } \left|\frac{x - t_j}{\Delta_m}\right| \leq 1,\\
0 & \mbox{otherwise}.
\end{cases}
\label{eq2}
\end{align}
The second equation in~(\ref{eq1}) that $Y_m$ should satisfy is often called interpolation condition.
An example of hat basis functions in arbitrary domain are shown in Fig.~\ref{hat}. Note that the $\phi_j (x)$ are bounded between $[0,1]$. Additionally, it satisfies the  equation: $\sum_{j=1}^{m}\phi_j (x_i) = 1$. This is the main reason why it is introduced to ease constrained GP problem. Other properties and applications of hat function can be referred in\cite{mirzaee2018application}.

\subsection{Algorithm for Hat-GP regression}
Now, let $\xi_j := Y(t_j)$ for $j =0, \cdots, m-1$. In GP framework, the conditional distribution is the variable that everyone cares about. To get the posterior mean and variance of output $y^{*}$ with test input $x^{*}$, we should compute the distribution of $Y_m$ conditionally on interpolation condition shown in equation~(\ref{eq1}). Observe that the vector $\xi$ at the knots is Gaussian vector with covariance matrix $\Gamma_{\theta}$: $\Gamma_{\theta}= (K(t_i,t_j))_{0 \leq i,j \leq m-1}$. For a more intuitive representation, we rewrite the interpolation condition in matrix form: $\Phi_x\xi + \varepsilon  \approx y$,  where $\Phi_x$ is the $n \times m$ matrix defined by $\Phi_x[i, j] = \phi_j (x_i)$, for all $i $$=$$0, \cdots, n-1$ and $j $$=$$0, \cdots, m-1$. This is obviously the general GP model, except that $f(x)$ is now approximated to $\Phi_x\xi$. Under this circumstance, the kernel matrix in Hat-GP is expressed as follows: $\Phi_x \Gamma_{\theta} \Phi_x^{T}$. To train Hat-GP, we only need to optimize the log marginal likelihood with respect to the introduced parameters. And it is very easy to extend to the case with multidimensional input training data. For example, in two-dimensional case, the $m_1 \times m_2$ numbers of knots should be introduced on a regular grid. Notice that each row of the matrix $\Phi_x$ now becomes:
\begin{equation}
\begin{split}
\Phi_x[i, ] = &[ \phi^{1}_{1}(x_1^i) \phi_{1}^2(x_2^i),  \cdots, \phi_{m_1}^1(x_1^i) \phi_{1}^2(x_2^i), \cdots,  \\
&\phi_{1}^1(x_1^i) \phi_{m_2}^2(x_2^i), \cdots, \phi_{m_1}^1(x_1^i) \phi_{m_2}^2(x_2^i) ],
\end{split}
\end{equation}
and the input for the matrix $\Gamma_{\theta}$ becomes two-dimensional knots points. 

The detailed Hat-GP algorithm is shown in algorithm~\ref{myalgorithm1}. Actually, it is the standard procedure to train Exact-GP model but with completely 
different mathematical expressions for the variables we care about. Certainly, MC and MCMC samplers can be performed to sample $\xi$ from the Gaussian distribution, which is not required to obey the inequality constraints in our case\cite{lopez2018finite, pakman2014exact}. These two methods will produce almost the same results. 

\begin{algorithm}[t]
	\caption{Algorithm for Hat-GP regression}\label{myalgorithm1}
	 \hspace*{\algorithmicindent} \textbf{Input: training data $(x, y)$, test input data $x^{*}$} \\
         \hspace*{\algorithmicindent} \textbf{Output: predicted mean and covariance value of test} \\
         \hspace*{\algorithmicindent} \textbf{\ \ \ \ output $y^{*}$: $\mu(x^{*})$, $\Sigma(x^{*})$}
	\begin{algorithmic}[1]
		\STATE Minimize negative log marginal likelihood with kernel parameters $\theta$, constant value $\mu_c$ and tandard deviation of noise 
		$\sigma$: 
		\STATE \begin{align}
			    &\frac{1}{2} (y-\Phi_{x}\mu_c)^\top (K_{xx}+  \sigma^{2}I)^{-1} (y-\Phi_{x}\mu_c)  \nonumber           \\
		            &+   \frac{1}{2} \log |K_{xx}+\sigma^{2}I | + \frac{n}{2} \log(2 \pi)       
		              \end{align}
		where $K_{xx}$$=$$\Phi_x \Gamma_{\theta} \Phi_x^{T}$.
		\STATE Compute the approximate conditional mean $\mu(x^{*})$ and covariance $\Sigma(x^{*})$  of $y^{*} |\{ \Phi_x \xi + \varepsilon \approx y \}$ with respect to the optimized 
		parameters ($\theta^{*}$, $\mu^{*}_c$ and $\sigma^{*}$): 
		\STATE
		          \begin{align}
                           &\Phi_{x^{*}}\mu^{*}_{c}+K_{x^{*}x} (\Phi_x \Gamma_{\theta^{*}} \Phi_x^{T} + \sigma^{*2} I)^{-1}(y-\Phi_{x}\mu^{*}_{c}) \\
			  &K_{x^{*}x^{*}}+ \sigma^{*2}I -K_{x^{*}x}(\Phi_x \Gamma_{\theta^{*}} \Phi_x^{T} + \sigma^{*2} I)^{-1} K_{xx^{*}} 
                            \end{align}
                where $K_{x^{*}x} = \Phi_{x^{*}} \Gamma_{\theta^{*}} \Phi_x^{T}$, $K_{xx^{*}} = \Phi_{x} \Gamma_{\theta^{*}} \Phi_{x^{*}}^{T}=K^{T}_{x^{*}x}$ and 
                             $K_{x^{*}x^{*}} = \Phi_{x^{*}} \Gamma_{\theta^{*}} \Phi_{x^{*}}^{T}$. 
	\end{algorithmic}
\end{algorithm}

\section{Experiments}
This section illustrates the validity of the proposed method. We test Hat-GP method on 1D and 2D datasets, and compare it with Exact-GP and FITC-GP method that are typically used in a similar setting. We start with one-dimensional Snelson dataset, and then provide comparison on two-dimensional dataset sampled from an analytical function.

\subsection{One-dimensional case}
\begin{figure}[!t]
\centering
\includegraphics[width=3in]{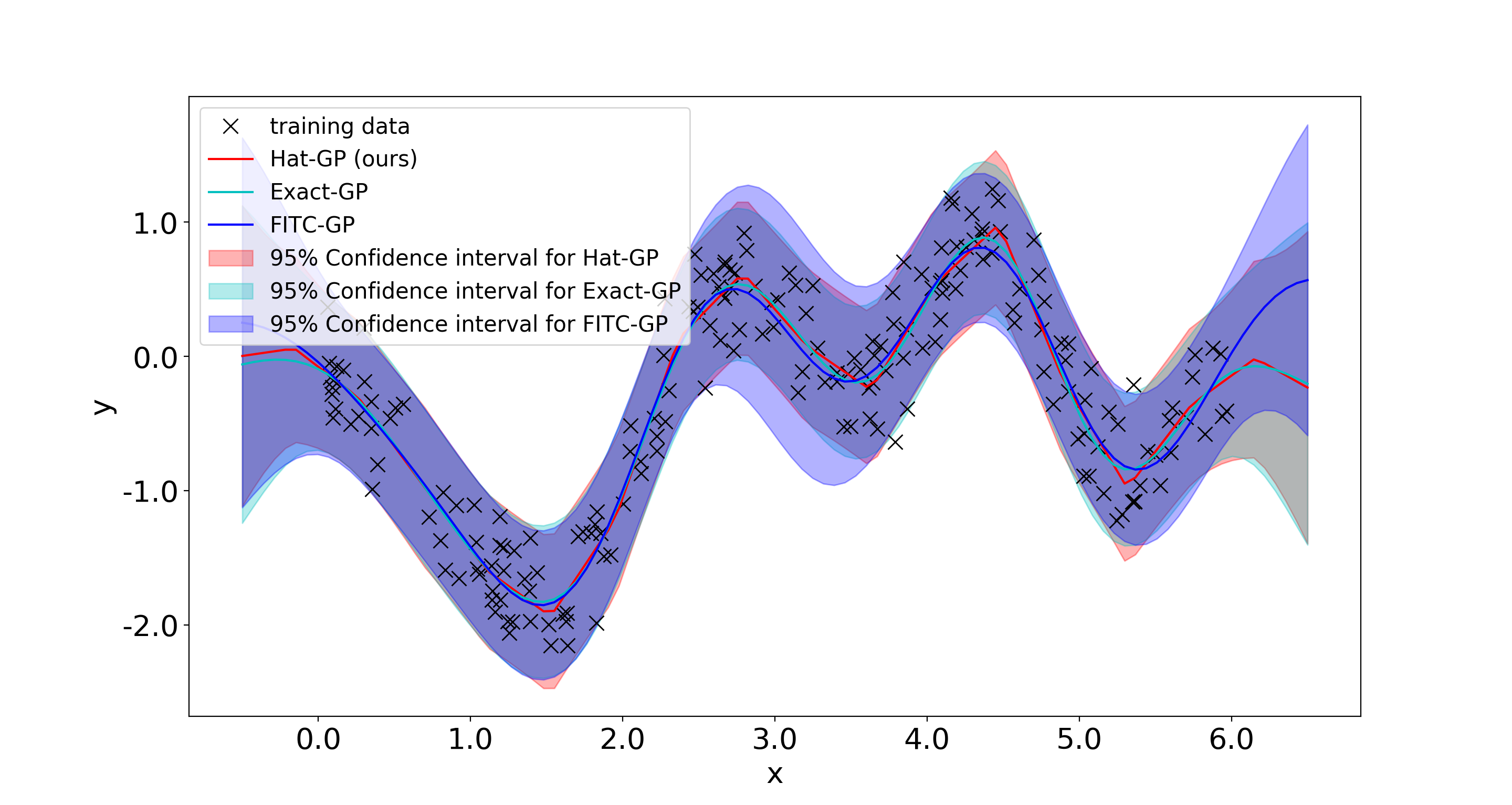}
\caption{The black crosses represent the training data. The predicted mean values of Hat-GP, Exact-GP, and FITC-GP are shown by red, cyan and blue lines, respectively. 
Their respective 95$\%$ confidence interval areas are represented by the same color.}
\label{hat2}
\end{figure}

In this case, we utilize the open-source data (Snelson dataset), which is downloaded from the URL link (\url{http://www.gatsby.ucl.ac.uk/~snelson/}), to test our method. SE kernel is used in this simulation. 20 numbers of hat basis functions are set in Hat-GP. In FITC-GP, 10 random pseudo points are spread in suitable domain for our simulation. And we do not optimize pseudo-inputs in this setting. The fitting results are shown in Fig.~\ref{hat2}. Training data are denoted as black cross. The predicted mean values of Hat-GP, Exact-GP and FITC-GP, along with their respective 95$\%$ confidence interval area, are shown in red, cyan and blue. The figure shows that the confidence intervals of Hat-GP and Exact-GP almost overlap. We can see that Hat-GP can achieve results comparable to the other two methods. If we increase the number of hat basis functions and pseudo points gradually, the red and blue lines will be getting smoother and continually approach the green line.

\subsection{Two-dimensional case}
\begin{figure}[!t]
\centering
\includegraphics[width=3in]{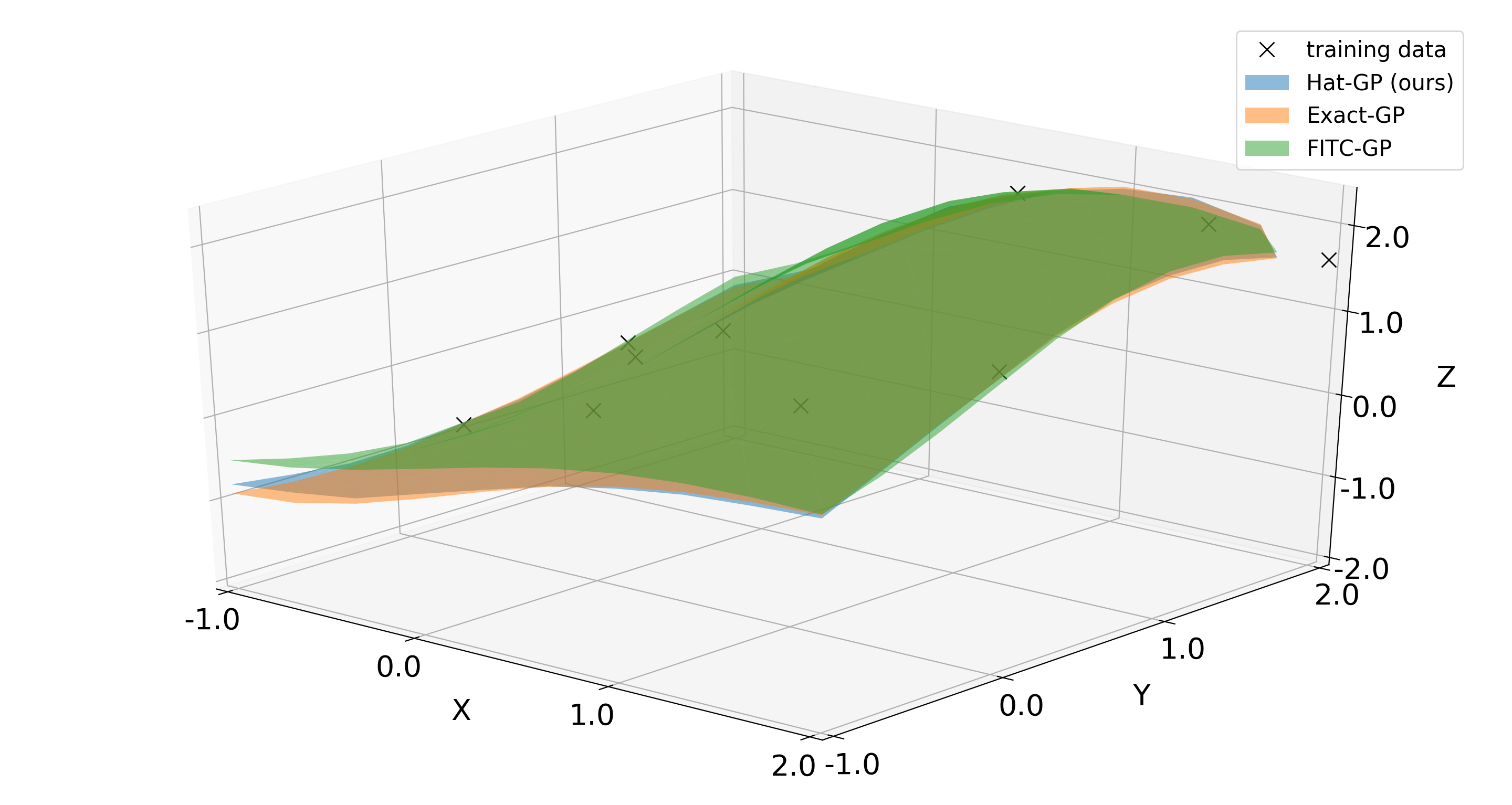}
\caption{The black crosses represent the training data. The predicted mean values of Hat-GP, Exact-GP and FITC-GP are shown with blue, orange and green surfaces, respectively.}
\label{hat3}
\end{figure}

In 2D toy example, 10 random training points are generated from 2D analytical function: $\arctan{5x} + \sin{1.5y}$. We use 2D ARD SE covariance function as our kernel. The number of hat basis functions in Hat-GP and pseudo-inputs in FITC-GP in each dimension are both 6. Similarly, pseudo-inputs are not optimized  in this simulation. The final recovery results are shown in Fig.~\ref{hat3}. The black crosses represent the training data. The predicted mean values of Hat-GP, Exact-GP and FITC-GP are shown with blue, orange and green surfaces, respectively.  The confidence interval areas are not shown for clarity. It illustrates that all these three methods can well capture the dominant features of the 2D analytical function. And we believe that if there are more training data, the surfaces will be much closer to the real analytical function.

\section{Discussions and Conclusions}
The Hat-GP method proposed in this paper is kind of spline interpolation put forward by Wahba\cite{wahba1990spline, wahba1978improper}. He proved that spline smoothing can be seen as GP regression with a specific choice of covariance function. Its convergent and asymptotic properties are discussed in detail in constrained GP and papers therein\cite{maatouk2017gaussian, lopez2018finite}.
It is a method closely related to the one of sparse GP methods, direct spectral approximations. The spectral analysis and series expansions of GP have a long history. Technically,  a continuous kernel functions  $K(x, x^{'})$ can be expanded into a Mercer series according to the Mercer's Theorem\cite{mercer1909xvi}:
$K(x, x^{'})=\sum _{ j=0 }^{ \infty  }\gamma_j \varphi_j(x)\varphi_j(x^{'})$,
where $\gamma_j$ and $\varphi_j$ are the eigenvalues and the orthonormal eigenfunctions of the covariance function. For realistic problem, we can approximate the kernel function by truncating the Mercer series and the approximation is guaranteed to converge to the exact covariance function when the number of terms is increased\cite{gorodetsky2016mercer, adler2010geometry}. 
In the case of GP, we can get that a nonzero-mean GP with Karhunen-Loeve series expansion\cite{ferrari1999finite, levy2008karhunen}: $f(x) \approx \sum_{ j=0 }^{J}f_{j} \varphi_j(x) $ ($J$ is the truncation number), where $f_j$ are independent nonzero-mean Gaussian random variables with variances $\gamma_j$. The generalization of this classical result to a continuum of eigenvalues was already discussed in Hilbert space methods for reduced-rank GP regression\cite{solin2020hilbert}. It approximates the Mercer expansion by using the basis consisting of the Laplacian eigenfunctions. However, in Hat-GP method, it is approximated by hat basis functions and the random sample $\xi_j$ in our case, which is still Gaussian random variables with variances $\Gamma_\theta$,  but $\Gamma_\theta$ is a non-diagonal matrix. So these Gaussian random samples are not independent. What's more, the hat basis functions are not orthonormal in its defined domain. These are the main differences between them. In addition, the inversion terms in algorithm~\ref{myalgorithm1} can be expanded according to the Woodbury formula, for example:
\begin{equation}
\begin{split}
(\Phi_x \Gamma_{\theta} \Phi_x^{T} +  \sigma^{2}I)^{-1} = \sigma^{-2}I- \sigma^{-4}\Phi_x(\Gamma^{-1}_{\theta}+\sigma^{-2}\Phi_x^{T} \Phi_x)^{-1}\Phi^{T}_x.
\end{split}
\end{equation}
Notice that $\Gamma^{-1}_{\theta}+\sigma^{-2}\Phi_x^{T} \Phi_x$ is a $m \times m$ matrix, which is usually much smaller than original $n\times n$ matrix before transformation.
After applying Woodbury formula, their computational complexity will reduce from $O(n^3)$ to $O(nm^2)$\cite{RW, solin2020hilbert}.
For efficient implementation, to avoid frequent matrix multiplication operations, these inversion terms are always carried out through Cholesky decomposition for numerical stability. 


In conclusion, we propose a novel sparse GP method based on hat basis functions. We extend the definition of hat basis functions mentioned in constrained GP and present a concise algorithm for regression problems. Its experimental performances on analytical functions or open-source datasets show that, like existing methods, it is a promising one for solving regression problems. 
In future work, we will conduct more extensive comparison between our method and several state-of-the-art methods to explore its scalability, efficiency and accuracy.  
We also hope to incorporate these methods into a unified framework.

\section*{Acknowledgment}
This work was supported by NSFC 61672512, U1632271, 61702493, Shenzhen S\&T Funding with Grant No. KQJSCX20170731163915914, CAS Key Laboratory of Human-Machine Intelligence-Synergy Systems, Shenzhen Institutes of Advanced Technology, and Shenzhen Engineering Laboratory for Autonomous Driving Technology.



%
\bibliographystyle{IEEEtran}
\bibliography{ICECCE}

\begin{thebibliography}{10}
\providecommand{\url}[1]{#1}
\csname url@samestyle\endcsname
\providecommand{\newblock}{\relax}
\providecommand{\bibinfo}[2]{#2}
\providecommand{\BIBentrySTDinterwordspacing}{\spaceskip=0pt\relax}
\providecommand{\BIBentryALTinterwordstretchfactor}{4}
\providecommand{\BIBentryALTinterwordspacing}{\spaceskip=\fontdimen2\font plus
\BIBentryALTinterwordstretchfactor\fontdimen3\font minus
  \fontdimen4\font\relax}
\providecommand{\BIBforeignlanguage}[2]{{%
\expandafter\ifx\csname l@#1\endcsname\relax
\typeout{** WARNING: IEEEtran.bst: No hyphenation pattern has been}%
\typeout{** loaded for the language `#1'. Using the pattern for}%
\typeout{** the default language instead.}%
\else
\language=\csname l@#1\endcsname
\fi
#2}}
\providecommand{\BIBdecl}{\relax}
\BIBdecl

\bibitem{RW}
C.~K. Williams and C.~E. Rasmussen, \emph{Gaussian processes for machine
  learning}.\hskip 1em plus 0.5em minus 0.4em\relax MIT press Cambridge, MA,
  2006, vol.~2, no.~3.

\bibitem{cardelli2019robustness}
L.~Cardelli, M.~Kwiatkowska, L.~Laurenti, and A.~Patane, ``Robustness
  guarantees for bayesian inference with gaussian processes,'' in
  \emph{Proceedings of the AAAI Conference on Artificial Intelligence},
  vol.~33, 2019, pp. 7759--7768.

\bibitem{overstall2020bayesian}
A.~M. Overstall, D.~C. Woods, and B.~M. Parker, ``Bayesian optimal design for
  ordinary differential equation models with application in biological
  science,'' \emph{Journal of the American Statistical Association}, pp. 1--16,
  2020.

\bibitem{yang2019multi}
K.~Yang, M.~Emmerich, A.~Deutz, and T.~B{\"a}ck, ``Multi-objective bayesian
  global optimization using expected hypervolume improvement gradient,''
  \emph{Swarm and evolutionary computation}, vol.~44, pp. 945--956, 2019.

\bibitem{iyer2019nonparametric}
K.~G. Iyer, E.~Gawiser, S.~M. Faber, H.~C. Ferguson, J.~Kartaltepe, A.~M.
  Koekemoer, C.~Pacifici, and R.~S. Somerville, ``Nonparametric star formation
  history reconstruction with gaussian processes. i. counting major episodes of
  star formation,'' \emph{The Astrophysical Journal}, vol. 879, no.~2, p. 116,
  2019.

\bibitem{jiang2019review}
Q.~Jiang, X.~Yan, and B.~Huang, ``Review and perspectives of data-driven
  distributed monitoring for industrial plant-wide processes,''
  \emph{Industrial \& Engineering Chemistry Research}, vol.~58, no.~29, pp.
  12\,899--12\,912, 2019.

\bibitem{solin2020hilbert}
A.~Solin and S.~S{\"a}rkk{\"a}, ``Hilbert space methods for reduced-rank
  gaussian process regression,'' \emph{Statistics and Computing}, vol.~30,
  no.~2, pp. 419--446, 2020.

\bibitem{bui2017unifying}
T.~D. Bui, J.~Yan, and R.~E. Turner, ``A unifying framework for gaussian
  process pseudo-point approximations using power expectation propagation,''
  \emph{The Journal of Machine Learning Research}, vol.~18, no.~1, pp.
  3649--3720, 2017.

\bibitem{cramer2013stationary}
H.~Cram{\'e}r and M.~R. Leadbetter, \emph{Stationary and related stochastic
  processes: Sample function properties and their applications}.\hskip 1em plus
  0.5em minus 0.4em\relax Courier Corporation, 2013.

\bibitem{wilson2015kernel}
A.~Wilson and H.~Nickisch, ``Kernel interpolation for scalable structured
  gaussian processes (kiss-gp),'' in \emph{International Conference on Machine
  Learning}, 2015, pp. 1775--1784.

\bibitem{maatouk2017gaussian}
H.~Maatouk and X.~Bay, ``Gaussian process emulators for computer experiments
  with inequality constraints,'' \emph{Mathematical Geosciences}, vol.~49,
  no.~5, pp. 557--582, 2017.

\bibitem{lopez2018finite}
A.~F. L{\'o}pez-Lopera, F.~Bachoc, N.~Durrande, and O.~Roustant,
  ``Finite-dimensional gaussian approximation with linear inequality
  constraints,'' \emph{SIAM/ASA Journal on Uncertainty Quantification}, vol.~6,
  no.~3, pp. 1224--1255, 2018.

\bibitem{duvenaud2014automatic}
D.~Duvenaud, ``Automatic model construction with gaussian processes,'' Ph.D.
  dissertation, University of Cambridge, 2014.

\bibitem{hartmann2019laplace}
M.~Hartmann and J.~Vanhatalo, ``Laplace approximation and natural gradient for
  gaussian process regression with heteroscedastic student-t model,''
  \emph{Statistics and Computing}, vol.~29, no.~4, pp. 753--773, 2019.

\bibitem{zhang2018advances}
C.~Zhang, J.~Butepage, H.~Kjellstrom, and S.~Mandt, ``Advances in variational
  inference,'' \emph{IEEE transactions on pattern analysis and machine
  intelligence}, 2018.

\bibitem{benton2019function}
G.~Benton, W.~J. Maddox, J.~Salkey, J.~Albinati, and A.~G. Wilson,
  ``Function-space distributions over kernels,'' in \emph{Advances in Neural
  Information Processing Systems}, 2019, pp. 14\,939--14\,950.

\bibitem{snelson2006sparse}
E.~Snelson and Z.~Ghahramani, ``Sparse gaussian processes using
  pseudo-inputs,'' in \emph{Advances in neural information processing systems},
  2006, pp. 1257--1264.

\bibitem{quinonero2005unifying}
J.~Qui{\~n}onero-Candela and C.~E. Rasmussen, ``A unifying view of sparse
  approximate gaussian process regression,'' \emph{Journal of Machine Learning
  Research}, vol.~6, no. Dec, pp. 1939--1959, 2005.

\bibitem{naish2008generalized}
A.~Naish-Guzman and S.~Holden, ``The generalized fitc approximation,'' in
  \emph{Advances in neural information processing systems}, 2008, pp.
  1057--1064.

\bibitem{mirzaee2018application}
F.~Mirzaee and N.~Samadyar, ``Application of hat basis functions for solving
  two-dimensional stochastic fractional integral equations,''
  \emph{Computational and Applied Mathematics}, vol.~37, no.~4, pp. 4899--4916,
  2018.

\bibitem{pakman2014exact}
A.~Pakman and L.~Paninski, ``Exact hamiltonian monte carlo for truncated
  multivariate gaussians,'' \emph{Journal of Computational and Graphical
  Statistics}, vol.~23, no.~2, pp. 518--542, 2014.

\bibitem{wahba1990spline}
W.~Grace, \emph{Spline models for observational data}.\hskip 1em plus 0.5em
  minus 0.4em\relax Siam, 1990, vol.~59.

\bibitem{wahba1978improper}
G.~Wahba, ``Improper priors, spline smoothing and the problem of guarding
  against model errors in regression,'' \emph{Journal of the Royal Statistical
  Society: Series B (Methodological)}, vol.~40, no.~3, pp. 364--372, 1978.

\bibitem{mercer1909xvi}
J.~Mercer, ``Functions of positive and negative type, and their connection the
  theory of integral equations,'' \emph{Philosophical transactions of the royal
  society of London. Series A, containing papers of a mathematical or physical
  character}, vol. 209, no. 441-458, pp. 415--446, 1909.

\bibitem{gorodetsky2016mercer}
A.~Gorodetsky and Y.~Marzouk, ``Mercer kernels and integrated variance
  experimental design: connections between gaussian process regression and
  polynomial approximation,'' \emph{SIAM/ASA Journal on Uncertainty
  Quantification}, vol.~4, no.~1, pp. 796--828, 2016.

\bibitem{adler2010geometry}
R.~J. Adler, \emph{The geometry of random fields}.\hskip 1em plus 0.5em minus
  0.4em\relax SIAM, 2010.

\bibitem{ferrari1999finite}
G.~Ferrari-Trecate, C.~K. Williams, and M.~Opper, ``Finite-dimensional
  approximation of gaussian processes,'' in \emph{Advances in neural
  information processing systems}, 1999, pp. 218--224.

\bibitem{levy2008karhunen}
B.~C. Levy, ``Karhunen loeve expansion of gaussian processes,'' in
  \emph{Principles of Signal Detection and Parameter Estimation}.\hskip 1em
  plus 0.5em minus 0.4em\relax Springer, 2008, pp. 1--47.

\end{thebibliography}

\end{document}